\documentclass{article}

\usepackage[nonatbib,preprint]{neurips_2020}

\usepackage[utf8]{inputenc} 
\usepackage[T1]{fontenc}    
\usepackage{hyperref}       
\usepackage{url}            
\usepackage{booktabs}       
\usepackage{amsfonts}       
\usepackage{nicefrac}       
\usepackage{microtype}      
\usepackage{graphicx}
\usepackage{subcaption}
\usepackage[ruled,vlined]{algorithm2e}
\usepackage{mathtools}
\usepackage{xcolor}
\usepackage{wrapfig}

\newcommand\abs[1]{\left|#1\right|}

\usepackage[backend=biber,style=ieee,citestyle=numeric,uniquename=init,mincitenames=1,maxcitenames=2,url=false,doi=false,isbn=false,date=year]{biblatex}

\DeclareFieldFormat{citehyperref}{%
  \DeclareFieldAlias{bibhyperref}{noformat}
  \bibhyperref{#1}}  
  
  \DeclareCiteCommand{\parencite}
  {\usebibmacro{prenote}}
  {\usebibmacro{citeindex}%
    [\printtext[citehyperref]{\usebibmacro{cite}}]}
  {\multicitedelim}
  {\usebibmacro{postnote}}

\bibliography{references}

\title{Geometric Prediction: \\ Moving Beyond Scalars}

\author{
  Raphael J. L. Townshend\\
  Department of Computer Science \\
  Stanford University\\
   \texttt{raphael@cs.stanford.edu} \\
 \And
  Brent Townshend \\
  Department of Bioengineering \\
  Stanford University\\
   \texttt{bst@stanford.edu} \\
 \And
  Stephan Eismann \\
  Department of Applied Physics \\
  Stanford University\\
   \texttt{seismann@stanford.edu} \\
    \And
 Ron O. Dror \\
 Department of Computer Science \\
  Stanford University\\
   \texttt{rondror@cs.stanford.edu} \\
}

\begin{document}

\maketitle

\begin{abstract}
Many quantities we are interested in predicting are geometric tensors; we refer to this class of problems as \emph{geometric prediction}.  Attempts to perform geometric prediction in real-world scenarios have been limited to approximating them through scalar predictions, leading to losses in data efficiency.  In this work, we demonstrate that equivariant networks have the capability to predict real-world geometric tensors without the need for such approximations.  We show the applicability of this method to the prediction of force fields and then propose a novel formulation of an important task, biomolecular structure refinement, as a geometric prediction problem, improving state-of-the-art structural candidates.  In both settings, we find that our equivariant network is able to generalize to unseen systems, despite having been trained on small sets of examples.  This novel and data-efficient ability to predict real-world geometric tensors opens the door to addressing many problems through the lens of geometric prediction, in areas such as 3D vision, robotics, and molecular and structural biology.
\end{abstract}

\section{Introduction}

What distinguishes a vector in three dimensions, such as a velocity, from a grouping of three arbitrary scalars, such as three independent speeds?  The answer lies in how these quantities react to geometric transformations.  The values of the velocity are dependent on a frame of reference, which for example can be rotated, whereas the values of the speeds are unchanged by such a transformation.  In other words, velocity is a \emph{geometric} vector, whereas the speeds are three individual scalars.  Scalars and geometric vectors are special cases of a more general-purpose object known as a geometric tensor.

%

%


When we refer to geometric tensors, we are not talking about generic multi-dimensional arrays, such as those in TensorFlow \cite{Abadi2016}.  Key to a geometric tensor's definition is its response to geometric transformations, such as rotations and translations.  Geometric tensors also have an order, corresponding to the number of indices required to select one of their values: scalars are zeroth-order, geometric vectors are first-order, and matrix quantities such as the moment of inertia tensor are second-order.

While an order $o$ tensor can be thought of as a multi-dimensional array of dimension $o$ in a given frame of reference (or basis), it must also transform in a specific manner when the basis is transformed.  Repeating our velocity example, if we rotated our basis, the velocity vector would also rotate.  Thus, three arbitrary numbers are not an instantiation of a tensor unless they are accompanied with a specific basis and a definition of how said numbers should change under geometric transformations.  For a more detailed and concise mathematic treatment see \textcite{Dullemond}.

We define geometric prediction as the class of prediction problems for geometric tensors beyond the 0-order scalar case.  Many quantities that arise in scientific problems and everyday life are not scalars, but geometric tensors.  First-order tensor examples include fluid flow, traffic flow, robot arm movement in path planning, and atmospheric circulation.  Higher-order tensors include the stress tensor and elasticity tensors in continuum mechanics (2nd and 4th order), the piezoelectric tensor in crystal study (3rd order), and the Riemann curvature tensor in differential geometry (4th order).

\begin{figure}
    \includegraphics[width=\textwidth]{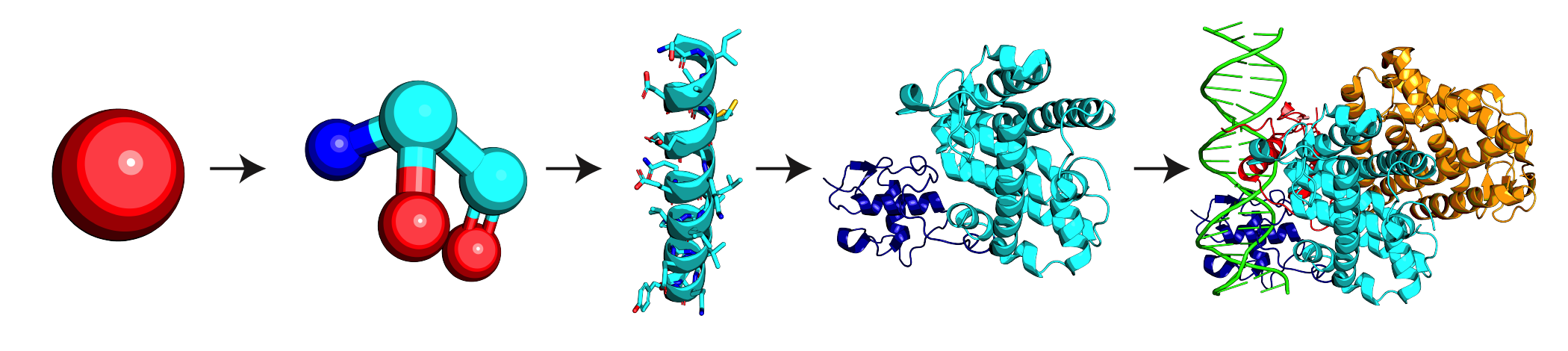}

  \caption{Hierarchy of geometric patterns.  Example biomolecular hierarchy shown, mapping from atoms to amino acids to secondary structure elements to domains to complexes.  Each level up involves the composition of subpatterns at different orientations and locations.  Equivariance allows for efficiently detecting and composing these geometric patterns at all levels of the hierarchy.}
   \label{fig:hierarchy}
\end{figure}

At the same time, commonly used neural network architectures do not have the ability to constrain their internal layers or their outputs to transform as a geometric tensor would.  And for every combination of lower-level patterns that create a useful higher-level pattern, this lack of constraints causes the network to have to relearn the correct combination for every possible geometric transformation (Figure \ref{fig:hierarchy}).  This is a strict and major loss of data efficiency when dealing with geometric tensors.  We are essentially performing structured prediction, with no actual enforcement of the structure. 

Recently, neural networks have emerged that do encode constraints such as rotations and translations \parencite{Cohen2016,Thomas2018b,Kondor2018f}.  These are known as equivariant networks, but to date their use has been restricted to internally representing geometric tensors, and they have not demonstrated the ability to predict geometric tensors in real-world scenarios.

In this work, we present three key contributions:

\begin{enumerate}
	\item We lay out the natural connection between geometric prediction and equivariant neural networks, opening the door to a number of novel applications for such techniques.
	\item We present the first real-world demonstrations of predicting non-scalar geometric tensors using equivariant networks.
	\item We formulate an important problem, biomolecular structure refinement, as a geometric prediction problem, implement an equivariant neural network, and achieve useful signal.
\end{enumerate}

In subsequent sections, we shall simply refer to geometric vectors and geometric tensors as vectors and tensors, respectively.

\section{Related Work}

Capturing core symmetries of data with neural networks has led to fundamental advances in machine learning, with convolutional neural networks \parencite{Krizhevsky2012} and graph neural networks \parencite{Duvenaud2015c} being among some of the most famous examples.  In a similar vein, many machine learning methods attempt to approximate these symmetries through data augmentation by including the same training examples in multiple poses, in areas such as astronomy \parencite{Dieleman2015} and medical imaging \parencite{Ronneberger2015}.  However, most learning methods do not try to predict tensors. 

In terms of geometric tensors, some works have modeled geometric predictions as sets of scalar predictions, employing a range of strategies.  In molecular force field prediction, where we wish to predict the forces that an atom will be subjected to, a recent example involved predicting each component of the resultant force vector as individual scalars, using data augmentation \parencite{Mailoa2019}.  In fluid dynamics, where we want to account for turbulence through the Reynolds stress tensor, said tensor was decomposed into a multiplication of a rotationally invariant neural network prediction and a fixed tensor operation \parencite{Ling2016}.  Again in fluid dynamics, another approach used Bayesian cluster neural networks to predict the velocity field resulting from the Navier-Stokes equations \parencite{McCracken2018}.  In all these approaches the neural networks in these approaches are all fundamentally predicting sets of individual scalars.

Another approach, in use for force field prediction of small systems \parencite{Chmiela2018, Xie2018, Zhang2017c}, is that of training a neural network to predict a scalar field, in this case the energy of a system, and then taking numerical gradients to get at the resultant vectors.  Many problems involving geometric tensors do not have an underlying scalar field, such as the stress tensor, magnetic fields, and optical flow vectors.  Even in cases where there is such a field, these methods require differentiation of each atom's prediction with respect to all positions within its receptive field, a computationally expensive step.  In this work, we consider tensors that are not necessarily derived from scalar fields, and extend to much larger systems than such methods have considered.

There has recently been significant excitement around equivariant networks that have the ability to reason about arbitrary tensors \parencite{Cohen2016,Worrall2017,Thomas2018b,Kondor2018f,Anderson2019,Weiler2018b}, and though they do not predict tensors, such methods have demonstrated significant results on scalar prediction problems in medical imaging \parencite{Winkels2019}, shape recognition \parencite{Esteves2019,Weiler2018b}, and property prediction for small molecules \parencite{Anderson2019} and macromolecules such as protein complexes \parencite{Eismann2020}. \textcite{Thomas2018b} presents the only demonstration of non-scalar predictions.  They use 2 toy problems involving less than 30 particles: acceleration and moment of inertia under Newtonian gravity, as well as inferring the location of a deleted atom.  Yet there has been no demonstration of using equivariant networks for the prediction of tensors in a real-world scenario.

\section{Equivariance}

A concept key to efficiently performing geometric prediction, as opposed to approximating such predictions with scalar predictors, is the notion of equivariance.  Intuitively, a function is equivariant to a transformation if such a transformation applied to its inputs can be instead expressed as a corresponding transformation of its output.  This allows a predictor's tensor output to correctly vary with geometric transformations of its input.

Formally, a function between vector spaces $f: X \rightarrow Y$ is equivariant to a group of transformations $G$ if we have:

$$\rho_Y(g) f(x) = f(\rho_X(g) x) \quad \forall x \in X \quad \forall g \in G$$

Where $\rho_X$ and $\rho_Y$ are group representations of $G$ in $X$ and $Y$, respectively.  A group representation $\rho$ is a mapping from a group element to an invertible square matrix that respects the property that $\rho(g_1 g_2) = \rho(g_1) \rho(g_2) \; \forall g_1, g_2 \in G$.  This square matrix has dimensions $n \times n$ where $n$ is the dimension of the underlying vector space.

As an example, take $X$ and $Y$ as both being the space of three-dimensional vectors, and $G = SO(3)$ being the space of rotations.  We could take both our group representations to be the mapping from a rotation to its corresponding $3 \times 3$ rotation matrix.  We would then say our function is equivariant to rotations (and the selected representations) if and only if a rotation of the input must lead to a rotation of the output.  

Invariance is a special case of equivariance where the output remains unchanged, i.e. $\rho_Y(g) = I$.  For example, most graph convolutional networks are invariant to rotations.  In many cases our desired output truly is invariant to rotations, and by building this constraint into our method increase our overall data efficiency.  In other words, if we encode symmetries present in the data into our model, we can decrease the model's variance at no cost to its bias.

While invariance is enough for scalars, we need methods to implement equivariance to be able to correctly transform the output of geometric tensors with transformations of the input signal.  Again, though we could attempt to output these quantities without equivariance, the removal of the corresponding constraints causes the learning problem to become significantly harder.  In fact, these difficulties only increase as the order of our geometric tensors and the dimension of our input space increases. 
  
Furthermore, many systems are composed of repeated patterns that are transformed in various ways with respect to one another and can be composed into higher-level patterns that depend on their subunit's transformations.  For example, proteins are composed of amino acids that can be rotated and translated in many ways, and are combined together to form larger units of structure known as secondary structure.  These secondary structure units depend on the orientation of translation of each of their component amino acids, and can themselves be transformed and composed into even larger groupings (see Figure \ref{fig:hierarchy}).

To be able to efficiently and accurately model these subunits and their transformations at all levels of the hierarchy, we need methods that can implement local equivariance in each of their layers.  A model such as a 3-dimensional convolutional neural network could in theory capture the details of the protein system described above, but would need to re-learn each pattern, and each useful combination of patterns, for every possible rotation, leading to an explosion in the data and computation requirements.  With equivariance, we maintain the ability to capture detailed geometry while also avoiding such costly overparameterization.

\section{Experiments}

\subsection{Network architecture}

\begin{wrapfigure}{r}{0.55 \textwidth}
    \includegraphics[width=0.55\textwidth]{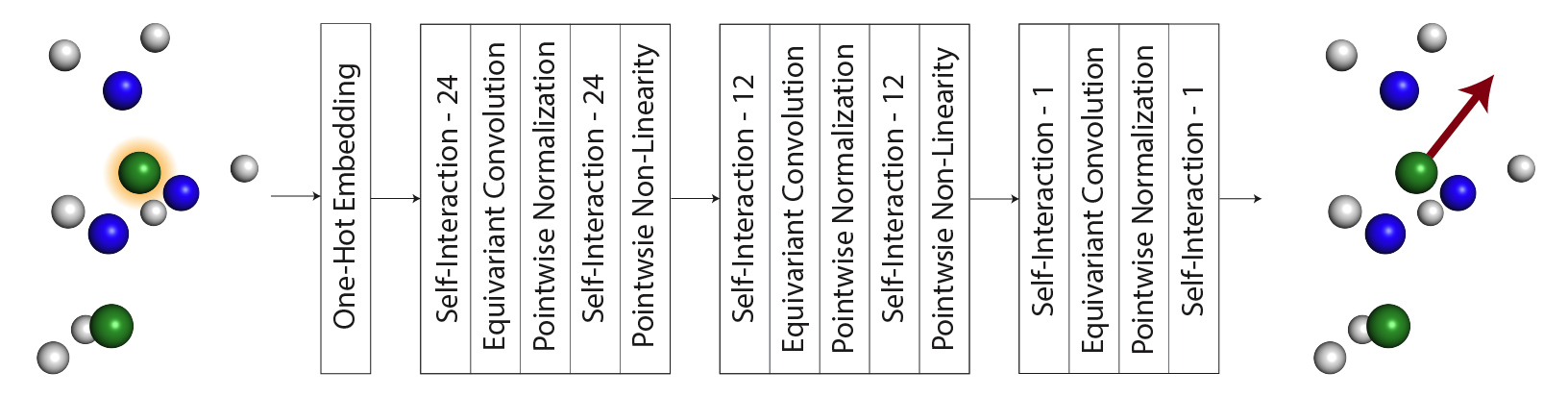}

  \caption{Model architecture.  We take in three-dimensional atomic coordinates and atom types around a target atom, and output a vector for that target atom.  Numbers indicate the number of filters in the self-interaction layers.  The entire network is equivariant to rotations and translations.}
   \label{fig:architecture}
\end{wrapfigure}

For both of our experiments, we create a model based on the layers defined in \parencite{Thomas2018b}, using 3 layers of equivariant convolutions with filter sizes of 24, 12, and 1 performed on each atom's 50 nearest neighbors.  For feature inputs we embed each element type $\{C, H, O, N, S\}$ into a one-hot encoding.  We use equivariant non-linearity and normalization layers after each convolution, as well as interleave 6 self-interaction layers (full model architecture in Figure \ref{fig:architecture}).  Our batches consist of all the atoms of one macromolecule at a time to ensure the network fit in GPU memory, and each network is trained over 75000 batches.  The memory and runtime of this network scales linearly with the number of particles in each system, as well as the total number of systems used.  For both tasks, we train 3 sets of models.  These sets only differ from each other in terms of the maximum tensor order that can be represented by their equivariant convolutions (orders 0, 1, and 2; internally these layers use spherical harmonics, and we achieve this by limiting their maximum angular frequency).  Each set contains 3 replicates, which vary only by the shuffling of their input examples within the training and validation sets, as well as the initialization of their weights.  Empirically, the memory and runtime scale quadratically with the maximum order represented.  All training and testing is performed on a single Titan X Pascal GPU, taking approximately 24 hours for the most expensive model. 

The 0-order case is essentially a standard graph convolutional network with no equivariance, with pairwise distances acting as edge features, and learnable node features.  It is invariant to rotations, and thus unable to predict tensor quantities, but we can use it as a comparison for predicted scalars.  

\subsection{Loss and metrics}
\label{metrics}

We optimize against a tensor version of the Huber loss \parencite{Huber1964}, $l$, defined as follows:

\[
  l(\mathbf{v}_p, \mathbf{v}_t) =
  \begin{cases}
                                   \frac{1}{2} {||\mathbf{v}_p - \mathbf{v}_t||}_2^2 & \text{if $||\mathbf{v}_p - \mathbf{v}_t|| \leq \delta$,} \\
                                   ({||\mathbf{v}_p - \mathbf{v}_t||}_2 - \frac{1}{2} \delta)\delta & \text{otherwise.}
  \end{cases}
\]

Where we are taking the $L_2$ norm of the difference between predicted tensors $\mathbf{v}_p$ and target tensor $\mathbf{v}_t$, and averaging over each batch.

For test-time evaluation, we select a range of metrics to provide a representative measure of our geometric predictions.  Specifically, we evaluate average tensor deviation, magnitude deviation, and angle (or magnitude-independent) deviation between our predicted and target tensors.  We cannot rely only on magnitude or angle metrics as they would not give a complete picture of the tensor prediction, while using only a tensor loss would not give us a detailed understanding of the relative contributions of the angle and magnitude errors.

\begin{enumerate}

\item Tensor --- magnitude of element-wise difference:
$$||\mathbf{v}_p - \mathbf{v}_t||_2$$ 
\item Magnitude --- absolute difference between magnitudes:
$$\abs{{||\mathbf{v}_p||}_2 - {||\mathbf{v}_t||}_2}$$ 
\item Angle --- angular similarity:
$$\arccos (\frac{\mathbf{v}_p \cdot \mathbf{v}_t}{{||\mathbf{v}_p||}_2 {||\mathbf{v}_t||}_2})$$

\end{enumerate}

We also evaluate characteristics of our overall distribution of predictions by examining the Pearson correlation of our predicted tensor magnitudes  ${||\mathbf{v}_p||}_2$ and the true tensor magnitudes ${||\mathbf{v}_t||}_2$.  As our order-0 model cannot predict tensors, we have it instead directly predict the scalar magnitude, using a standard scalar Huber loss, and report only performance on the magnitude metric.  Finally, note that while our current prediction targets are first-order tensors (geometric vectors), internal layers can use higher-order tensors and the above metrics are applicable to any order tensor.

\subsection{Molecular force fields}

We begin by predicting an important physical quantity: the force applied on an atom at a given point in time.  This quantity is the product of an atom's interactions with nearby atoms, with the strength of these interactions determined by proximity.  These forces are used in molecular simulations, where we are interested in how molecules change shape over time and how they react to external stimuli.  They must be re-computed at an extremely high frequency (every 2 femtoseconds of simulation time is common), and thus accelerated approximations could increase the speed of such simulations.

The training and testing data are produced by mining the Protein Data Bank \parencite{Berman2000} for structures containing 1) protein 2) a single chain 3) less than 200 amino acids 4) better than 2.5 Å resolution, and 5) solved by X-ray crystallography.  We further split the data based on sequence identity of 20\%, using the software BLASTP \parencite{Altschul1990a} to avoid similar structures overlapping in the train, validation, and test sets.  For each of these sets we randomly sample 100, 50, and 50 proteins, respectively.  We compute the forces that would be applied on a biomolecular structure using the OpenMM software package \parencite{Eastman2017}, using the AMBER14 force field \parencite{Maier2015} with explicit solvent.  Note this is a classical force field that approximates the underlying quantum mechanical forces.  We keep the solvent atoms for computing our predictions, but only predict against the protein atoms' computed forces.  We set the Huber loss parameter $\delta$ to 10 meV/Å and select a learning rate of 0.1 based on manual hyperparameter search on the validation set, with an order-1 model.  We also performed an experiment where we doubled the number of filters in each layer, finding no significant effect.  A total of 31 models were trained for this task.  

\begin{figure}
  \centering
  \begin{subfigure}{0.32\textwidth}
    \includegraphics[width=\textwidth]{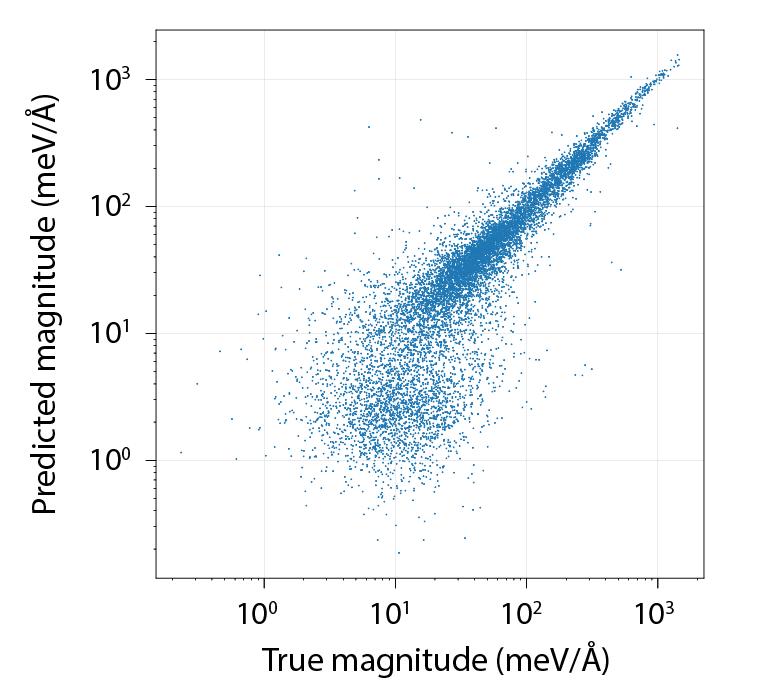}
  \end{subfigure}
  \begin{subfigure}{0.32\textwidth}
    \includegraphics[width=\textwidth]{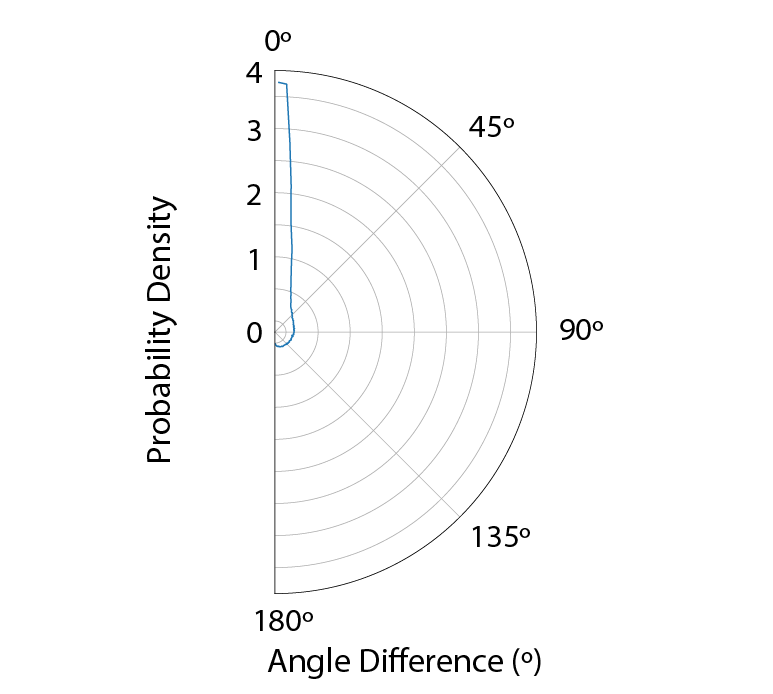}
  \end{subfigure}
    \begin{subfigure}{0.32\textwidth}
    \includegraphics[width=\textwidth]{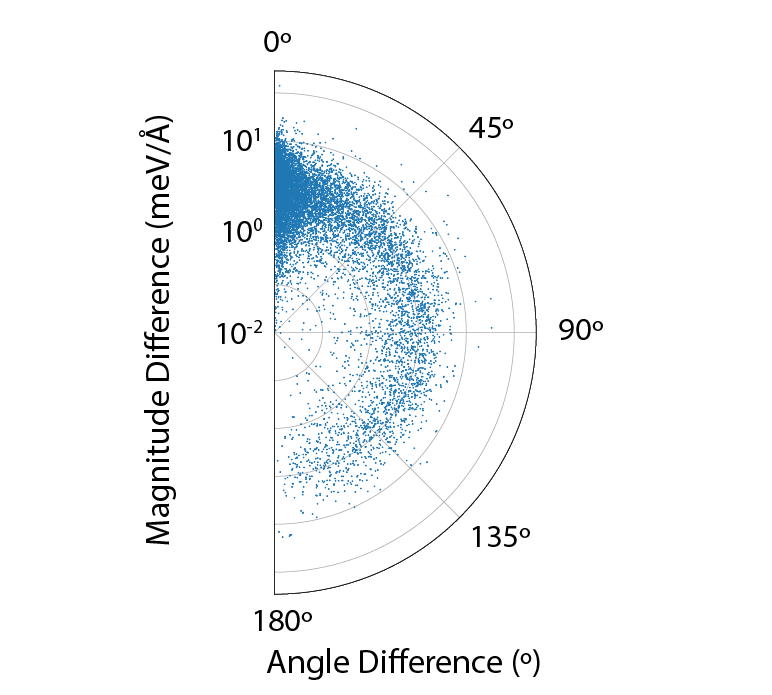}
  \end{subfigure}
  \caption{Force field prediction performance of the best model, as selected by validation loss; order 2.  We compare true and predicted magnitudes (left panel), angular similarity histogram (middle panel), and the angular similarity versus the absolute difference between magnitudes (right panel).  Angular similarities are displayed in polar coordinates, with 0º meaning no angular difference.}
\label{fig:ff}
\end{figure}

Looking at the distribution of predictions for our best predictor (order 2, selected by validation loss) in Figure \ref{fig:ff}, we observe strong overall linear correlation (0.98 Pearson) between the predicted and true magnitudes at all distances, indicating that our network is able to account for both detailed local perturbations and much larger-scale forces.  The distribution of the magnitude of the errors remain constant across different target prediction magnitudes, implying the errors are additive.  We also observe that the vast majority of predictions have little angular deviation from the correct direction, which is consistent with the tensor error not being significantly worse than the magnitude error.

As seen in Table \ref{table:ff}, we obtain better performance compared to naive predictors in all three of our metrics for models that use equivariance.  While the order-0 predictor (a standard graph convolutional network with no equivariance) does not obtain useful signal in predicting the magnitude, we see that the addition of orientation information that comes with the order-1 predictor leads not only to the ability to predict the angular and tensor terms, but also to a dramatic improvement in the magnitude prediction.  The order-2 model improves on all metrics even further, showing that the ability to model higher-order tensors internally aids overall performance.  The tensor error is, as expected, higher than the magnitude error as it must also include angle errors in its computation, highlighting the fact that we are not simply re-predicting the magnitude of the underlying vectors, but are in fact outputting geometric vectors that approximate the correct values.

\begin{table}
\begin{center}
  \begin{tabular}{  l@{\qquad} c@{\qquad} c@{\qquad} c@{\qquad}  c  }
  \toprule
    Metric & Naive & Order 0 & Order 1 & Order 2 \\ 
    \midrule
    Magnitude (meV/Å) &  79.9   & 77.8 (79.4 $\pm$ 1.0) & 23.0 (22.9 $\pm$ 0.1)  & \textbf{15.6 (17.4 $\pm$ 1.0)} \\
    Angle (degrees)   & 90.0 & ---   &  42.4 (43.5 $\pm$ 0.6) & \textbf{32.8 (33.8 $\pm$ 0.6)} \\
    Tensor (meV/Å)      & 81.5   & ---   & 28.3 (28.5 $\pm$ 0.1)   & \textbf{20.7 (22.4 $\pm$ 0.9)} \\
    \bottomrule
  \end{tabular}
\end{center}
\caption{Force field prediction metrics.  We predict the forces computed by the AMBER force field.  We examine our predictions through magnitude-, angle-, and tensor-based metrics.  Each metric compares the predicted tensor to the true tensor, with the magnitude metric being the mean absolute error of the magnitudes, the angle metric being mean angular similarity, and the tensor metric being the mean of the magnitude of the difference (see Section \ref{metrics} for details).  Our naive models are the mean magnitude prediction, a random direction, and an all zero prediction, respectively.  The order-0 model is a standard graph convolutional network that is invariant to rotation.  We report the best model (by val loss), as well as the mean and standard error of the mean across three trained models.}
\label{table:ff}
\end{table}


The closest prior results we identified that predict atomic fields is \textcite{Mailoa2019}.  They predicted force fields for three systems: a polymer and two small molecules.  They trained and tested each model on the same molecule using various timepoints in a molecular dynamics simulation, using data augmentation to account for rotations. They reported an average relative error of their predictions of 0.13, 0.31, and 0.31 for the three systems. We compute a relative tensor error for our selected order-2 model: $\frac{||\mathbf{v}_p - \mathbf{v}_t||_2}{||\mathbf{v}_t||_2} = 0.25$ on our test set, which was a distinct set of molecules from any in training and included larger extended systems.


We hypothesize that our network's ability to generalize to new structures, while using only 100 training structures, stems from the data efficiency that comes with treating our prediction targets as tensors, not sets of scalars.  Overall, these results demonstrated the network's ability to learn a complex closed-form geometric function with high-accuracy. 

\subsection{Structure Refinement}

We next apply this model to biomolecular structure refinement, posing the task as a geometric prediction problem.  Structure refinement is a common task in structural biology that involves perturbing a hypothetical structure so that its geometric arrangement more closely resembles the structure it actually adopts.  Using physical force fields to move the atoms into their appropriate locations usually requires a prohibitively expensive number of steps, and due to the local alignments there is no continuously differentiable scalar field to take gradients against to determine how to move the atoms.  Note that local alignment is necessary since small changes at one point in a long molecule can result in large changes in global alignment of distant atoms.

\begin{figure}
\centering
    \includegraphics[width=0.8 \textwidth]{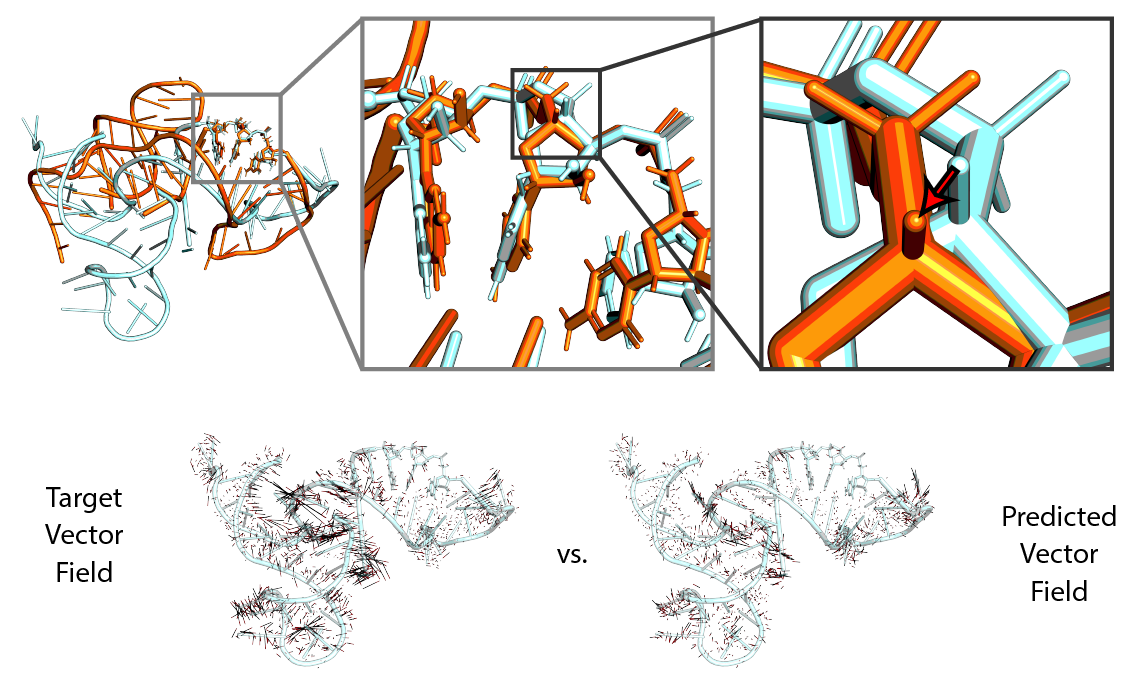}

  \caption{Structure refinement vector field visualized for a candidate structure (cyan) with respect to the ground truth structure (orange).  While the full structure can have large global deviations (top left panel), we can be robust to those and focus on structure refinement by aligning the local environment on an atom (top middle panel).  For this atom, this yields a refinement vector (top right panel), which is our prediction target.  If we repeat this process for every atom in the candidate structure, we end up with our structure refinement vector field (bottom left panel) which we can then try and predict (bottom right panel).  See Algorithm \ref{alg:srvf} for a more formal definition.}
   \label{fig:sp-example}
\end{figure}

We instead define a local deviation tensor that points in the direction to move a candidate structure's atoms to locally reduce its deviation from the ground truth (Figure \ref{fig:sp-example}, Algorithm \ref{alg:srvf}).  To the best of our knowledge, there have been no attempts to predict such a quantity before.

\begin{algorithm}
\DontPrintSemicolon
\SetAlgoLined
\SetKwInOut{Input}{Input}
\SetKwInOut{Output}{Output}
\Input{Atoms $\mathcal{A}$; current positions $\mathcal{P}^c = \{\mathbf{p}_a^c, \forall a \in \mathcal{A}\}$; target positions $\mathcal{P}^t = \{\mathbf{p}_a^t, \forall a \in \mathcal{A}\}$; nearest neighbor function $\mathcal{N}: a \times \mathcal{P} \rightarrow 2^\mathcal{A}$; alignment function $\mathcal{L} : \mathcal{P}_{mobile} \times \mathcal{P}_{target} \rightarrow (\mathcal{T}: \mathbf{p} \rightarrow \mathbf{p})$}
\Output{Refinement vectors $\{ \mathbf{v}_a$, $\forall a \in \mathcal{A} \}$}
 \For{$a \in \mathcal{A}$}{
 $\mathcal{P}_a^c \leftarrow \{\mathbf{p}_i^c,\forall i \in \mathcal{N}(a, \mathcal{P}^c)\}$ \tcp*{Get nearest atoms in candidate structure.}
 $\mathcal{P}_a^t \leftarrow \{\mathbf{p}_i^t, \forall i \in \mathcal{N}(a, \mathcal{P}^c)\}$ \tcp*{Look up corresponding atoms in native.}
 $\mathcal{T}_a \leftarrow L(\mathcal{P}_a^t, \mathcal{P}_a^c)$  \tcp*{Align the native atoms to the candidate atoms.}
 $\mathbf{v}_a \leftarrow \mathbf{p}_a^c - \mathcal{T}(\mathbf{p}_a^t)$ \tcp*{Use resulting transformation to get target vector.}
  }
 \caption{Structure Refinement Vector Field}
 \label{alg:srvf}
\end{algorithm}

We generate data for our structure refinement task by using the structural candidates for each of 21 RNA sequences provided in the FARFAR2-Puzzles set \parencite{Watkins2019}.  Each sequence was the target of a blind computational challenge in the RNA Puzzles series \parencite{Cruz2012}, and comes with an experimentally determined ground truth structure, as well as many computationally generated candidate structures that aim to approximate this ground truth.  We split these in the order that the true structures were experimentally determined, using the first 13 sequences as training, the next 4 as validation, and the last 4 as testing.  As these structural candidates are the product of state-of-the-art RNA structure prediction software, improving upon them represents a difficult but important challenge.

We generate our structure refinement tensors for each candidate structure, using the corresponding ground truth structure as the target.  We use a learning rate of 0.1 selected via manual hyperparameter tuning on the validation set.  We set $\delta = 1$ Å.  A total of 136 models were trained for this task. 

\begin{figure}
  \centering
  \begin{subfigure}{0.32\textwidth}
    \includegraphics[width=\textwidth]{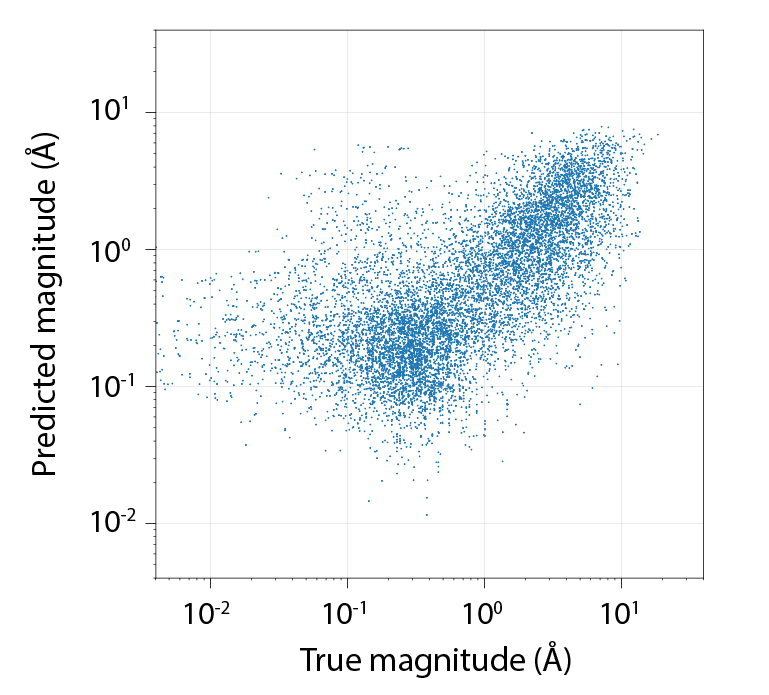}
  \end{subfigure}
  \begin{subfigure}{0.32\textwidth}
    \includegraphics[width=\textwidth]{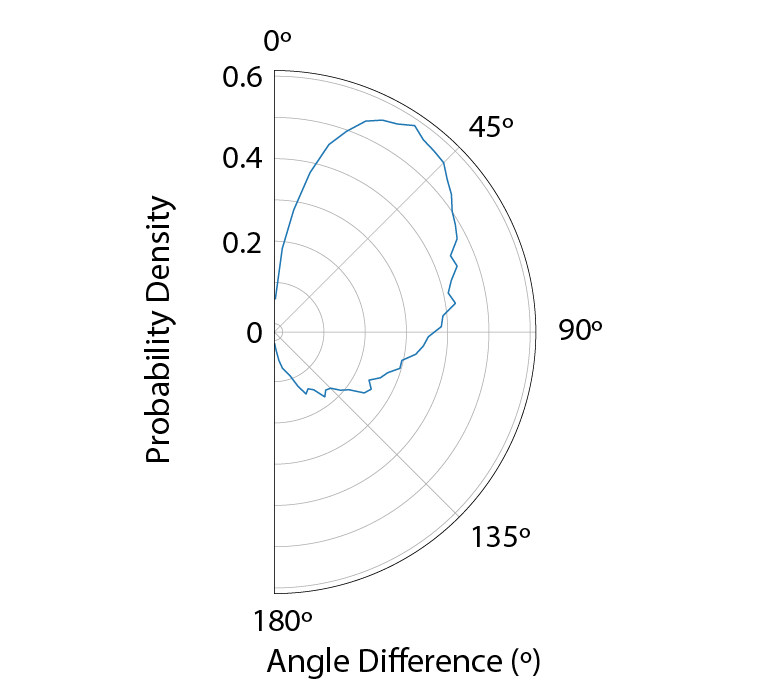}
  \end{subfigure}
     \begin{subfigure}{0.32\textwidth}
    \includegraphics[width=\textwidth]{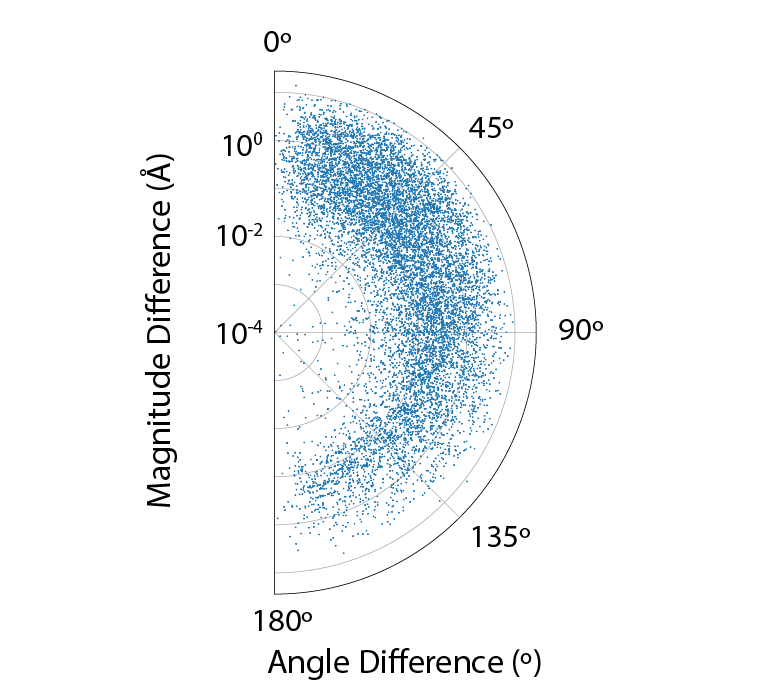}
  \end{subfigure}
  \caption{Local structure refinement performance of the best model, as selected by validation loss; order 2.  We show the same distributions as in Figure \ref{fig:ff}.  }
\label{fig:sp}
\end{figure}

Our results, shown graphically in Figure \ref{fig:sp}, indicate that we are able to improve the candidate structures, opening the door to improving the local structure refinement of a state-of-the-art RNA prediction software.  For magnitudes (left panel) we obtain a Pearson correlation of 0.66.  The errors in this case appear to scale with the magnitude of the target vector, reflecting a multiplicative relation between the target magnitude and the error.  For angles (middle panel) we see significant enrichment for correct angle predictions though less than in the molecular force field prediction case, likely due to the difficulty of the structure refinement problem.  In our angular versus magnitude error comparison, we observe a fairly uniform spread of magnitude error across different angular errors.

The vector predictions of the order-2 model for one example are shown in Figure \ref{fig:sp-example} bottom-right, overlaid on the candidate structure.  This illustrates that \ regions of the candidate structure that are far from the ground truth are also the regions that the network predicts have large target vectors.

The metrics in Table \ref{table:sp} show that increasing the order of tensors our model is able to represent leads again to improved performance, though contrasted to molecular force field prediction, the angular predictions are noisier, reflecting the difficulty of the structure refinement problem.  This difficulty also appears in the difference between the tensor and the magnitude errors.  

Our network is again able to generalize to unseen structures, improving on the results of a state-of-the-art structure predictor, even though trained on only 13 RNA structures.  Overall, these results demonstrate our network is able to predict vectors on a difficult problem with no closed-form solution.


\begin{table}
\begin{center}
  \begin{tabular}{  l@{\qquad} c@{\qquad}  c@{\qquad} c@{\qquad} c  }
  \toprule
    Metric & Naive & Order 0 & Order 1 & Order 2   \\ 
    \midrule
    Magnitude (Å) & 1.55 & 1.38 (1.38 $\pm$ 0.03) & 1.03 (1.02 $\pm$ 0.01) &  \textbf{0.95 (0.99 $\pm$ 0.02)} \\
    Angle (Degrees) & 90.0 &  --- & 71.1 (70.8 $\pm$ 0.1) & \textbf{68.8 (69.8 $\pm$ 0.5)}\\
    Tensor (Å) & 1.60 & --- & \textbf{1.46 (1.49 $\pm$ 0.01)} &  \textbf{1.45 (1.48 $\pm$ 0.02)}\\
    \bottomrule
  \end{tabular}
\end{center}
\caption{Local structure refinement metrics.  We use the same metrics and baselines as in Table \ref{table:ff}.}
\label{table:sp}
\end{table}

\section{Conclusion}

In this work, we defined geometric prediction as the prediction of tensors---a class of problems that to date has largely been modeled without leveraging their inherent structure.  We discussed equivariance's key role in geometric prediction, in terms of increasing data efficiency and allowing for detection and composition, under geometric transformation, of local elements.  

We demonstrated that an equivariant network was able to predict molecular force fields, an important physical quantity.  We also formulated an important problem, biomolecular structure refinement, as a geometric prediction problem, and achieved significant predictive power, using an equivariant network, in improving candidate structures generated by state-of-the-art RNA structure prediction software.  These results also highlight the data efficiency of this equivariant network, as it was able to perform well despite being trained on only small sets of data.

While these results are promising, we anticipate that there remain substantial areas for improvement.  The design of equivariant networks, the definition of further geometric prediction problems, and the derivation of novel loss functions for such tasks all emerge as fruitful directions for further research.

Taken together, we hope to spur further interest in the development of algorithms for geometric prediction, as well as the formulation of further geometric prediction problems.  We expect such prediction problems to be relevant in most areas involving geometric systems, such as structural and molecular biology, robotics, and computer vision. 

\section*{Broader Impact}

Our work can be applied to a broad range of applications that are associated with geometric relationships to increase fidelity, reduce bias, and reduce training data requirements of systems.  In terms of broader impacts, here we focus on the impact on applications related to modeling of the structure of biomolecules, an area in which we have also used for characterizing the performance of our work.

The shape of biomolecules in a large part dictates their functional characteristics. Improvements in the ability to predict structure of biomolecules directly lead to better understanding of these molecules, which in turn can be applied to human health in terms of diagnostics and therapies. In the longer term, this understanding will improve our ability to prevent, diagnose, and treat disease in humans, animals, and plants. 

Inherent biases in any ML system can result in erroneous or misinterpreted results. When applied to medical decision-making or treatment, the results of such errors can be devastating. The complexity of these systems, including the present work, make identification of the causes of such errors difficult to identify, resulting in uncertainty as to whether a system is performing correctly.  However, application of symmetries that are physically based, such as the equivariance enforced in our models, result in a reduction of training requirements and in the complexity needed of a network to perform a given task;  this should also reduce the propensity of unanticipated results.

Use of particular benchmarks, such as the PDB databases used in this work, may lead to systems and models that are skewed in their performance toward these benchmarks, resulting in lower performance in other cases, and possibly on systems to which they are ultimately applied. Before deployment or reliance on these systems, especially in the medical arena, audits of any biases and the match between training data and the ultimate applications should be carefully assessed.

\section*{Acknowledgments}

The authors thank Alex Derry, Milind Jagota, Bowen Jing, Masha Karelina, Yianni Laloudakis, Joe Paggi, Alex Powers, Patricia Suriana, and Martin Vogele for their discussions and advice. This work was supported by Intel, the U.S. Department of Energy Office of Science Graduate Student Research (SCGSR) program (R.T.), a Stanford Bio-X Bowes fellowship (S.E.), and the U.S. Department of Energy, Scientific Discovery through Advanced Computing (SciDAC) program (R.D.).

\newpage

\printbibliography

\end{document}